\documentclass[sigconf]{acmart}

\usepackage{graphicx}
\usepackage{booktabs}
\usepackage{array}
\usepackage{adjustbox}
\usepackage{xspace}
\usepackage{multirow} 
\usepackage{enumitem} 
\usepackage{xcolor}
\usepackage{bm}
\usepackage{algorithm}
\usepackage[noend]{algpseudocode}
\usepackage{amsmath}
\usepackage{setspace} 
\usepackage{balance}

\AtBeginDocument{%
  }

\copyrightyear{2025}
\acmYear{2025}
\setcopyright{cc}
\setcctype{by-nc}
\acmConference[CIKM '25]{Proceedings of the 34th ACM International
Conference on Information and Knowledge Management}{November 10--14,
2025}{Seoul, Republic of Korea}
\acmBooktitle{Proceedings of the 34th ACM International Conference on
Information and Knowledge Management (CIKM '25), November 10--14, 2025,
Seoul, Republic of Korea}\acmDOI{10.1145/3746252.3761085}
\acmISBN{979-8-4007-2040-6/2025/11}

\begin{document}
\title{ST-LINK: Spatially-Aware Large Language Models for Spatio-Temporal Forecasting}

\author{Hyotaek Jeon}
\affiliation{%
  \institution{Pohang University of Science and Technology}
  \city{Pohang}
  \country{Republic of Korea}}
\email{taek98@postech.ac.kr}

\author{Hyunwook Lee}
\affiliation{%
  \institution{Pohang University of Science and Technology}
  \city{Pohang}
  \country{Republic of Korea}}
\email{hwlee0916@postech.ac.kr}

\author{Juwon Kim}
\affiliation{%
  \institution{Pohang University of Science and Technology}
  \city{Pohang}
  \country{Republic of Korea}}
\email{juwona@postech.ac.kr}

\author{Sungahn	Ko}
\authornote{Corresponding Author}
\affiliation{%
  \institution{Pohang University of Science and Technology}
  \city{Pohang}
  \country{Republic of Korea}}
\email{sungahn@postech.ac.kr}

\newcommand{\toolname}{ST-LINK\xspace}
\newcommand{\memory}{MRFFN\xspace}
\newcommand{\sako}[1]{\textcolor{red}{[#1 -SK-]}}

\begin{abstract}
Traffic forecasting represents a crucial problem within intelligent transportation systems. In recent research, Large Language Models (LLMs) have emerged as a promising method, but their intrinsic design, tailored primarily for sequential token processing, introduces notable challenges in effectively capturing spatial dependencies. Specifically, the inherent limitations of LLMs in modeling spatial relationships and their architectural incompatibility with graph-structured spatial data remain largely unaddressed.
To overcome these limitations, we introduce ST-LINK, a novel framework that enhances the capability of Large Language Models to capture spatio-temporal dependencies. Its key components are Spatially-Enhanced Attention (SE-Attention) and the Memory Retrieval Feed-Forward Network (MRFFN). SE-Attention extends rotary position embeddings to integrate spatial correlations as direct rotational transformations within the attention mechanism. This approach maximizes spatial learning while preserving the LLM's inherent sequential processing structure. Meanwhile, MRFFN dynamically retrieves and utilizes key historical patterns to capture complex temporal dependencies and improve the stability of long-term forecasting. 
Comprehensive experiments on benchmark datasets demonstrate that ST-LINK surpasses conventional deep learning and LLM approaches, and effectively captures both regular traffic patterns and abrupt changes.
\end{abstract}

\ccsdesc[500]{Information systems~Spatial-temporal systems}
\ccsdesc[500]{Computing methodologies~Neural networks}

\keywords{Large Language Models; LLM; Spatio-Temporal Forecasting; Urban Computing}

\maketitle

\section{Introduction}
Spatio-temporal forecasting is one of the well-studied forecasting problems, which models an intrinsic spatio-temporal interdependency. 
In contrast to traditional time-series forecasting, which focuses on modeling temporal relationships, spatio-temporal forecasting is established on the domains where the spatial dependencies are indispensable. 
For example, in traffic forecasting, a representative example of spatio-temporal modeling, certain road's traffic condition highly depends on spatially related roads (e.g., neighboring) and such dependency varies over time and events (e.g., accidents or traffic signals).
Effectively encoding both spatial and temporal features remains a key challenge in developing accurate spatio-temporal forecasting.

To model these spatio-temporal dependencies, various deep learning techniques have been proposed. Early work includes recurrent architectures like DCRNN~\cite{dcrnn} and methods applying Graph Convolutional Networks (GCNs) to encode road network connectivity~\cite{stgcn,astgcn,astgnn}. However, GCN-based models often rely on predefined graphs, which limits their adaptability to dynamic spatial structures. To overcome this, attention-based models~\cite{gman,STTN} began replacing static graphs with spatial attention for more flexible modeling. Transformer-based architectures~\cite{pdformer, traffictransformer, staeformer}, in particular, have been explored for their ability to capture long-range dependencies.

As transformer-based models prove their effectiveness in long-range dependency modeling, larger transformer-based models--commonly referred to as \textit{foundation model}--have emerged as a pivotal research theme in spatio-temporal forecasting~\cite{traffic_survey}. Initial efforts included training BERT-based models on large-scale traffic data~\cite{trafficbert} or leveraging structured prompts to encode spatial relationships~\cite{urbangpt}. More recently, to improve spatial learning, researchers have integrated deep learning-based embeddings with LLMs. For instance, GATGPT~\cite{gatgpt} integrates Graph Attention Networks, TPLLM~\cite{tpllm} leverages CNN and GCN-based features, and ST-LLM~\cite{stllm} incorporates embeddings that explicitly encode location and temporal patterns.

While extensive research has explored the integration of LLMs into spatio-temporal forecasting by incorporating spatial information into pre-trained models, existing LLM-based approaches primarily rely on introducing additional embedding before LLM input or conducting prompt engineering. To fully leverage the potential of foundation models and develop a outstanding framework for spatio-temporal forecasting, it is essential to introduce a novel methodology that preserves the architectural strengths of LLMs as well as effectively integrates intrinsic spatial and temporal dependencies. In this paper, we introduce ST-LINK,  a novel methods that enhances LLMs' ability to process spatio-temporal data while maintaining their architectural efficiency. ST-LINK consists of two techniques: Spatially-Enhanced Attention (SE-Attention) and Memory Retrieval Feed-Forward Network (MRFFN), focusing on improving LLMs’ ability to capture spatial dependencies.

\begin{figure}[t] 
    \centering
    \includegraphics[width=\columnwidth]{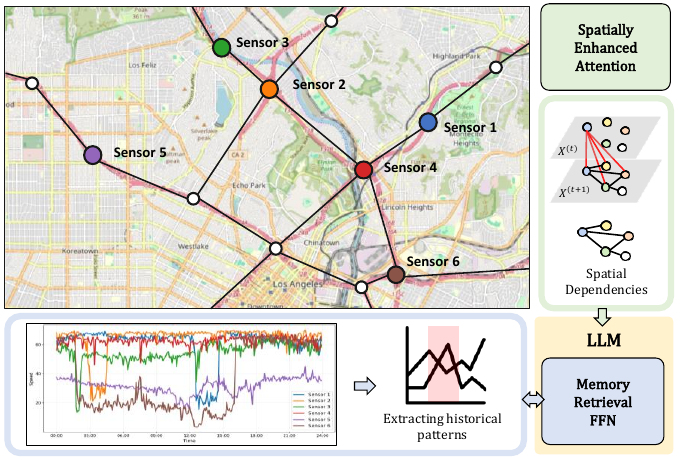} 
    \caption{Overview of the ST-LINK, illustrating its architecture with Spatially-Enhanced Attention(SE-Attention) and Memory Retrieval FFN(MRFFN), and its application to spatio-temporal data.}
    \label{fig:framework_overview}
\end{figure}

To effectively integrate and extend LLM's ability into spatial dimension, we first introduce SE-Attention, which extends Rotary Positional Embedding (RoPE) \cite{roformer} into a two-dimensional representation that explicitly encodes spatial relationships.
By extending RoPE to spatial dimension, ST-LINK\footnote{Github Link: \url{http://github.com/HyoTaek98/ST_LINK}} ensures a unified representation of spatio-temporal dependencies, preserving both spatial and temporal structures. 
In addition, instead of relying solely on the sequential order of tokens (i.e., temporal dependencies), our method decomposes query and key embeddings into separate temporal and spatial components, applying two-dimensional rotary transformations to align with the underlying spatio-temporal structure. Introducing SE-Attention module for each LLM block makes LLM more effectively adapt to the spatio-temporal forecasting task by establishing better spatio-temporal dependency modeling. 


Furthermore, to support the long-range dependency modeling, we design MRFFN to mitigate the limitation of forecasting models, especially in long-term forecasting~\cite{pmmemnet, megacrn, testam}. 
Inspired by retrieval-augmented generation (RAG) \cite{rag_paper} in LLMs and memory-based spatio-temporal learning~\cite{testam,megacrn,pmmemnet}, MRFFN integrates a dynamic memory module that stores and retrieves key historical representations. 
Unlike existing methods that rely on static memory embeddings or external retrieval sources, our framework employs end-to-end memory learning, enabling adaptive recall of past patterns.
This mechanism allows the model to dynamically update and retrieve relevant historical patterns, preserving robust long-term dependencies.
Continuously refining memory states, MRFFN ensures that the model remains responsive to evolving spatio-temporal patterns.

The effectiveness of ST-LINK is validated through extensive experiments on traffic flow and demand forecasting benchmarks, demonstrating its ability to outperform traditional graph-based neural models and Transformer-based spatio-temporal architectures. Compared to existing LLM-based methods, ST-LINK enhances spatial learning, stabilizes long-term predictions. 
The main contributions of this work include that :
\begin{itemize}[leftmargin=*]
\item We identify key challenges in adapting LLMs for spatio-temporal forecasting, 
\item We propose \toolname, which consists of SE-Attention and MRFFN for effective spatio-temporal modeling, 
\item We provide extensive experiments and anlaysis study with existing forecasting models and LLMs, demonstrating the potential of LLMs in spatio-temporal forecasting. 
\end{itemize}


\section{Related Work}
\textbf{Spatio-Temporal Forecasting Models. } Traditional spatio temporal forecasting methods~\cite{arima,ha} primarily relied on statistical models to capture temporal patterns. While these approaches offer simplicity and interpretability, they struggle with non-linear characteristics and spatial dependencies in data. To overcome these limitations, deep learning-based models have been widely adopted. 
DCRNN \cite{dcrnn} introduced a recurrent architecture that models spatial transitions through a diffusion process, while STGCN \cite{stgcn} applied graph convolutional networks (GCNs) to encode road network connectivity using an adjacency matrix. However, GCN-based models suffer from over-smoothing, leading to information loss across nodes and difficulty in capturing long-range dependencies. To address this, Graph-Wavenet \cite{gwn} introduced an adaptive adjacency matrix with dynamic node embeddings, improving spatial dependency learning.  

More recently, attention-based methods have emerged as an alternative to fixed graph structures. GMAN \cite{gman}, PDFormer \cite{pdformer} replace adjacency matrices with self-attention mechanisms, allowing the model to dynamically infer spatial correlations. Additionally, various transformer-based approaches have been proposed to incorporate spatio-temporal learning. Similarly, STAEFormer \cite{staeformer} incorporates Spatio-Temporal Adaptive Embeddings with a vanilla Transformer, and STTN \cite{STTN} employs a spatial transformer to capture directed spatio-temporal dependencies.
To further improve forecasting performance, studies such as PM-MemNet \cite{pmmemnet}, MegaCRN \cite{megacrn}, and TESTAM \cite{testam} have employed memory-based architectures to enhance spatio-temporal modeling by retrieving relevant historical patterns. These approaches effectively strengthen the model’s ability to retain long-term dependencies and adapt to dynamic spatio-temporal variations.
\\
\textbf{Large Language Models (LLMs). } 
As LLMs continue to demonstrate strong performance across various domains, their application has expanded into spatio-temporal forecasting \cite{traffic_survey}. An early study, TrafficBERT \cite{trafficbert}, proposed a BERT-based model specialized for traffic data, leveraging large-scale traffic datasets for pretraining to assess its applicability across road networks. Subsequently, methodologies integrating prompt-based learning and deep learning-based spatio-temporal embedding techniques have been introduced to extend LLMs for spatio-temporal modeling. Notably, UrbanGPT \cite{urbangpt} adopts Spatio-Temporal Instruction tuning, utilizing structured prompts to explicitly encode spatial relationships. However, since prompt-based approaches rely on providing additional contextual information rather than enabling the model to inherently learn spatial dependencies, they exhibit high sensitivity to prompt formulation and suffer from limitations in generalization. 

Accordingly, researchers have increasingly focused on integrating LLMs with deep learning-based spatio-temporal embedding techniques to more effectively capture spatio-temporal dependencies. GATGPT \cite{gatgpt} combines Graph Attention Networks (GATs) with LLMs to enhance spatial dependency modeling, while TPLLM \cite{tpllm} employs CNN-based sequence embeddings and GCN-based spatial feature extraction to enable LLMs to learn structured representations. ST-LLM \cite{stllm} introduces spatio-temporal embeddings that directly encode location information and temporal patterns, facilitating the effective learning of spatial relationships by LLMs.

Although these approaches enhance spatial learning, they mainly adapt deep learning embeddings rather than developing methods tailored for LLM architectures. This misalignment with LLMs' tokenization and sequential processing hinders effective spatial dependency modeling and comprehensive spatio-temporal representation. 
To address these limitations, we propose \toolname, which aligns an LLM's sequential processing with spatio-temporal structures using an extended RoPE for spatial encoding and an MRFFN for pattern storage, enhancing the modeling of spatio-temporal dependencies.

\begin{figure*}[t] 
    \centering
    \includegraphics[width=\textwidth]{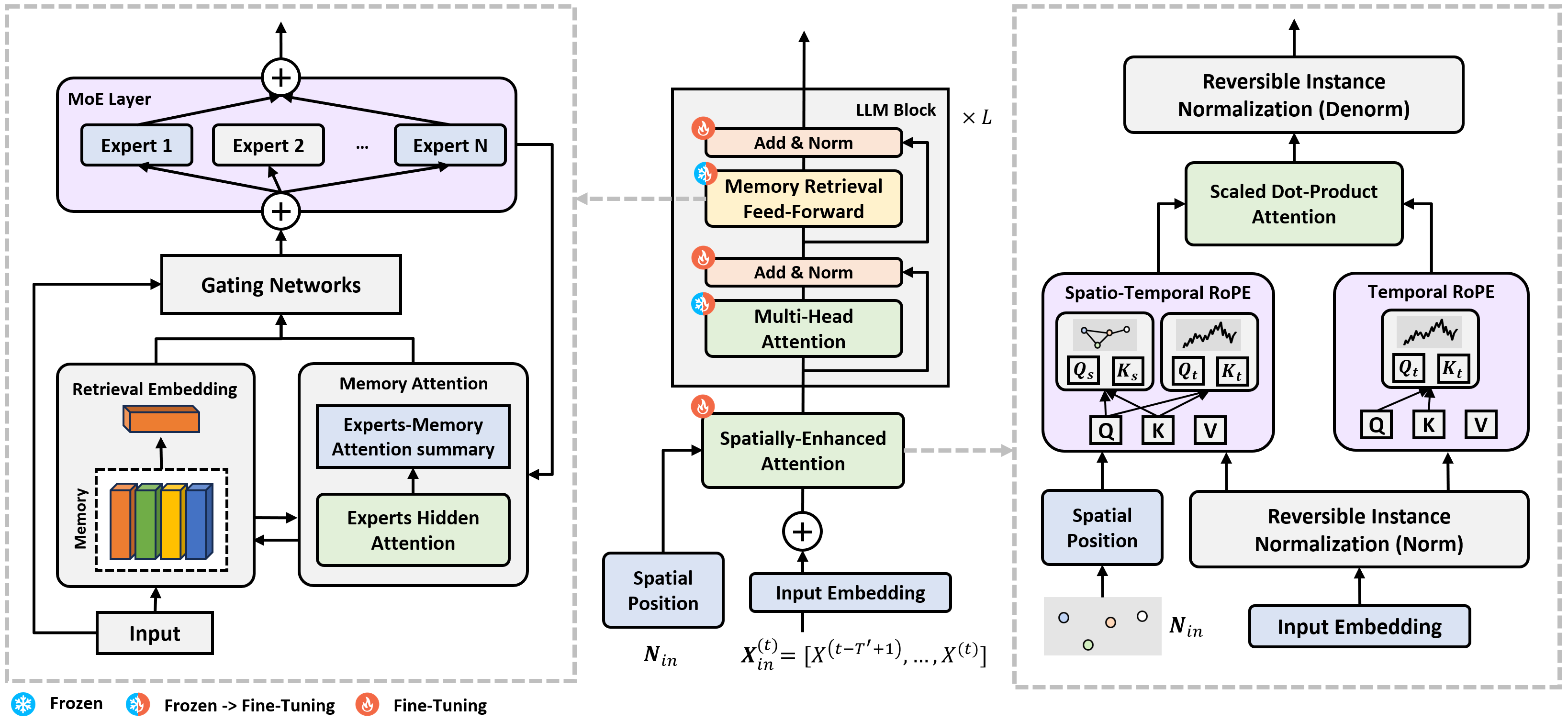} 
    \caption{Overview of the proposed \toolname. Left: MRFFN retrieves and integrates past spatio-temporal patterns for long-term forecasting. Middle: The LLM block and overall processing pipeline. Right: SE-Attention extends RoPE to encode spatial dependencies while preserving the LLM’s inherent temporal structure.}
    \label{fig:framework}
\end{figure*}
\section{Methodology}

\subsection{Preliminary}
\textbf{Problem Definition. } Spatio-temporal forecasting involves predicting future graph signals while considering both temporal dependencies and spatial correlations within a given network structure. Let us define a road network as \(\mathcal{G} = (\mathcal{V}, \mathcal{E}, \mathcal{A})\), where $\mathcal{V}$ represents the set of road segments with $|\mathcal{V}| = N$, $\mathcal{E}$ denotes the edges that define connectivity between roads, and $\mathcal{A} \in \mathbb{R}^{N \times N}$ is a matrix that encodes the topological structure of $\mathcal{G}$. The weights in this matrix are typically computed using empirical methods, such as a Gaussian kernel~\cite{shuman2013emerging} and cosine similarity~\cite{geng2019spatiotemporal} based on the road distance between sensor. Given the historical observations \(\mathbf{X}_{t-T^\prime+1}, \dots, \mathbf{X}_t \in \mathbb{R}^{N \times C}\) over the past $T^\prime$ time steps, the objective is to predict future graph signals \(\mathbf{Y}_{t+1}, \dots, \mathbf{Y}_{t+T} \in \mathbb{R}^{N \times C}\) for the next $T$ time steps while preserving the spatial relationships defined in $\mathcal{G}$. This forecasting task can be formally expressed as:
\begin{equation}
[\mathbf{X}_{t-T^\prime+1}, \dots, \mathbf{X}_t; \mathcal{G}] \xrightarrow[\theta]{f(\cdot)} [\mathbf{X}_{t+1}, \dots, \mathbf{X}_{t+T}]
\end{equation}
where $\theta$ represents the learnable parameters of the function $f(\cdot)$, and the mapping is defined as \(f: \mathbb{R}^{T^\prime \times N \times C} \rightarrow \mathbb{R}^{T \times N \times C}\)
Here, $C$ denotes the number of input and output features, such as traffic speed or demand volume. The function $f(\cdot)$ is designed to effectively capture both temporal dependencies and spatial correlations by leveraging past graph signals and the structural relationships defined in $\mathcal{G}$. 
\\ 
\\
\textbf{Embedding Methods. } To apply pre-trained large language models (LLMs) to various spatio-temporal forecasting tasks, it is necessary to introduce appropriate learning strategies or redesign input embedding layers to accommodate the unique characteristics of such data \cite{GPT4TS, timellm, llm4ts}. Traditional Transformer-based models typically employ Positional Embedding (i.e., PE) to encode absolute positional information~\cite{GPT1,BERT}. However, PE solely represents absolute positions and lacks mechanisms to capture relative positional relationships, making it less effective for modeling long-range dependencies. Furthermore, PE does not inherently encode structural relationships in spatio-temporal data, limiting its ability to capture spatial dependencies between nodes.  

Recently, rotary position embedding (i.e. RoPE)~\cite{roformer} has been introduced in various LLM architectures (e.g., LLAMA, Vicuna, Falcon) as an alternative to traditional PE \cite{llama2, falcon, vicuna}. 
Unlike PE, RoPE applies rotations to a query vector $\bm{q}$ and a key vector $\bm{k}$ using using rotation matrices, $\bm{R}_{m}$ and $\bm{R}_{n}$, which encode their absolute positions $m$ and $n$ respectively.
This formulation ensures that the resulting attention scores depend on relative positions, enabling effective modeling of long-range dependencies. 
This application of rotations ensures that the dot product in self-attention naturally depends on the relative positions of tokens ($(\bm{R}_m \bm{q})^T (\bm{R}_n \bm{k}) = \bm{q}^T \bm{R}_{m-n} \bm{k}$), where $\bm{R}_{m-n}$ represents the rotation matrix corresponding to the relative position $m-n$.
This property allows RoPE to provide more fine-grained positional representations, enhancing a model's ability to process long sequences efficiently.

The key idea of RoPE is to treat a $d$-dimensional vector (e.g., query $\bm{q}_p$ or key $\bm{k}_p$ at an absolute position $p$) as $d/2$ pairs of features. Each $i$-th pair is rotated in its 2D plane by an angle $\theta_{p,i}$. This rotation for a feature pair $[\bm{x_{2i}}, \bm{x_{2i+1}}]^T$ within such a vector is expressed as:
\begin{equation}
\begin{bmatrix} \bm{x'_{2i}} \\ \bm{x'_{2i+1}} \end{bmatrix}_p = \begin{bmatrix} \cos \theta_{p,i} & -\sin \theta_{p,i} \\ \sin \theta_{p,i} & \cos \theta_{p,i} \end{bmatrix} \begin{bmatrix} \bm{x_{2i}} \\ \bm{x_{2i+1}} \end{bmatrix}_p \label{eq:rope_pair_rotation}
\end{equation}
The rotation angle $\theta_{p,i}$ is determined by the absolute position $p$ and the pair index $i$:
\begin{equation}
\theta_{p,i} = p \cdot \omega_i, \quad \text{where} \quad \omega_i = \frac{1}{10000^{2i/d}}, \quad i \in [0, \dots, d/2-1] \label{eq:rope_angle_definition}
\end{equation}
Here, $\omega_i$ sets the rotation frequency for each feature pair. This formulation allows RoPE to capture positional dependencies in a structured manner across different dimensions.

\subsection{Model Architecture}
\subsubsection{\textbf{Spatially-Enhanced Attention}}
Traditional RoPE is primarily designed for capturing sequential relationships in textual data, focusing on temporal dependencies. However, applying standard RoPE directly to spatio-temporal data presents inherent limitations. It lacks mechanisms to explicitly model spatial dependencies between nodes or the complex network structures often found in spatio-temporal datasets. Consequently, while RoPE excels at encoding sequential patterns, it is not inherently suited for handling spatio-temporal dependencies, such as those required in traffic forecasting or demand forecasting tasks. 

To overcome these limitations, this study proposes an extended RoPE that explicitly incorporates spatial relationships within spatio-temporal data. The proposed approach preserves RoPE’s capability to capture temporal patterns while dynamically integrating spatial information, enabling LLMs to achieve superior performance in spatio-temporal learning tasks. 

SE-Attention is introduced to extend RoPE for spatial encoding while mitigating node-specific scale variations using Reversible Instance Normalization (RevIN)~\cite{revin}. Spatio-temporal data often exhibit significant variations in scale and distribution across different nodes, which can hinder model stability and predictive accuracy. To address this issue, we apply RevIN before the attention mechanism. RevIN performs instance-wise normalization for each node, which helps to standardize input feature distributions while preserving essential temporal patterns. A detailed description of RevIN is provided in Appendix~\ref{sec:revin}.

SE-Attention applies distinct rotary embeddings to the query $\bm{q_i}$ and key $\bm{k_i}$ vectors to encode temporal and spatial positional information. Each token $i$ is associated with a temporal position $t_i$ and spatial node $node_i$, and represented by $d$-dimensional vectors $\bm{q_i}, \bm{k_i} \in \mathbb{R}^d$ per attention head. Initially, standard temporal RoPE, denoted $\operatorname{RoPE}_{T}$, is applied to $\bm{q_i}$ and $\bm{k_i}$ using temporal position $t_i$ to rotate each feature pair, thus embedding temporal context within these vectors. Concurrently, spatial information is encoded by applying a specialized spatial RoPE variant $\operatorname{RoPE}_{S}$. For this spatial encoding, each node $node_i$ possesses a learnable spatial position embedding $N_{in}^{(node_i)}$. This node-specific embedding allows each spatial position to learn a unique scaling factor for the base RoPE frequencies, thereby generating distinct rotational patterns tailored for different nodes. Using this embedding, spatial rotation angles $\theta_{S,p}$ are computed from base RoPE frequencies $\omega_p = 1/10000^{2p/d}$, where $p \in [0, \dots, d/2-1]$ indexes the feature pairs:
\begin{equation}
\theta_{S,p}^{(node_i)} = N_{in}^{(node_i)} \cdot \omega_p
\label{eq:se_attn_spatial_angles_final_v2}
\end{equation}
The $\operatorname{RoPE}_{S}$ module processes an input vector $\bm{X} \in \{\bm{q_i}, \bm{k_i}\}$ by selectively applying spatial rotations. Specifically, only the initial $d/2$ dimensions of the input vector $\bm{X}$ are modified by these rotations, while the subsequent $d/2$ dimensions are preserved without alteration. Therefore, the resulting $d$-dimensional output from $\operatorname{RoPE}_{S}$ consists of this spatially rotated first half concatenated with the original, unchanged second half of $\bm{X}$. The rotation for each $p$-th feature pair $(X_{2p}, X_{2p+1})$ within the aforementioned initial $d/2$ dimensions (where $p$ indexes these $d/4$ pairs) using the angle $\theta_{S,p}^{(node_i)}$ is detailed as:
\begin{equation}
\begin{bmatrix} X'_{2p} \\ X'_{2p+1} \end{bmatrix} = 
\begin{bmatrix} \cos \theta_{S,p}^{(node_i)} & -\sin \theta_{S,p}^{(node_i)} \\ \sin \theta_{S,p}^{(node_i)} & \cos \theta_{S,p}^{(node_i)} \end{bmatrix}
\begin{bmatrix} X_{2p} \\ X_{2p+1} \end{bmatrix}
\label{eq:se_attn_spatial_transform_final_v2}
\end{equation}

After obtaining RoPE$_T$ and RoPE$_S$, the final query and key representations, $\Phi(\bm{q_i})$ and $\Phi(\bm{k_j})$, are formed by concatenating the outputs from $\operatorname{RoPE}_{T}$ and $\operatorname{RoPE}_{S}$. Subsequently, linear projections $\bm{W_q}$ and $\bm{W_k}$ reduce these combined $2d$-dimensional vectors back to their original dimension $d$.
\begin{equation}
\Phi(\bm{q_i}) = \bm{W_q} 
\begin{bmatrix} 
\operatorname{RoPE}_{T}(\bm{q_{i}}; t_i) \\ 
\operatorname{RoPE}_{S}(\bm{q_{i}}; N_{in}) 
\end{bmatrix},\ 
\Phi(\bm{k_j}) = \bm{W_k} 
\begin{bmatrix} 
\operatorname{RoPE}_{T}(\bm{k_{j}}; t_j) \\ 
\operatorname{RoPE}_{S}(\bm{k_{j}}; N_{in}) 
\end{bmatrix}
\end{equation}

The final attention output for the target token $i$ is computed using scaled dot-product attention as follows:
\begin{equation}
\text{Attention}(\bm{Q}, \bm{K}, \bm{V})_i = 
\sum_{j=1}^{\bm{N}} \frac{
\exp\left(\Phi(\bm{q_i})^\top \Phi(\bm{k_j}) / \sqrt{\bm{d_h}} \right)
}{
\sum_{n=1}^{\bm{N}} \exp\left(\Phi(\bm{q_i})^\top \Phi(\bm{k_n}) / \sqrt{\bm{d_h}} \right)
} \bm{v_j},
\end{equation}
where $N$ denotes the number of source tokens, and $d_h$ represents the head dimension. This approach ensures that the attention scores effectively integrate both spatial and temporal positional relationships. Finally, the output is denormalized using RevIN to restore the original scale.

\subsubsection{\textbf{Memory Retrieval Feed-Forward Network}}
Inspired by memory-based spatio-temporal models \cite{pmmemnet, megacrn, testam} and RAG \cite{rag_paper,rag_survey1}, which improves model adaptability by incorporating retrieved knowledge, we propose a novel Memory Retrieval Feed-forward Network (MRFFN) designed specifically for LLM architectures. 
MRFFN introduces an End-to-End Memory Learning mechanism that enables the model to store, retrieve, and update spatio-temporal patterns, improving long-term forecasting stability. To effectively incorporate memory-based learning, MRFFN integrates a Mixture-of-Experts (MoE)~\cite{moe,moe2,moe3}, allowing dynamic selection of specialized experts based on retrieved historical patterns.

MRFFN's memory module comprises a set of learnable key-value pairs, denoted as $\bm{k_m}, \bm{v_m}$. For an input $\bm{x} \in R^d$, where $d$ represents the model's hidden dimension, relevant historical patterns are retrieved. First, the top-$k$ memory slots are selected based on the similarity between $x$ and the memory keys $\bm{k_m}$. Then, attention weights $w_m$ are computed using softmax-normalized similarity scores, and these weights are used to aggregate the corresponding value vectors $\bm{v_m}$ into a retrieved embedding $\bm{z_r}$. This retrieval mechanism is formalized as:
\begin{equation}
    \bm{z_r} = \sum_{m \in \text{top-}k} w_m \bm{v_m}, \quad w_m = \frac{\exp(\bm{x} \cdot \bm{k_m})}{\sum_{m' \in \text{top-}k} \exp(\bm{x} \cdot \bm{k_{m'}})}
    \label{eq:mrffn_retrieval}
\end{equation}
The retrieved embedding $\bm{z_r}$ encapsulates input-influenced historical context. To ensure the memory module remains adaptive, MRFFN continuously refines its memory keys $\bm{k_m}$ using a momentum-based Exponential Moving Average (EMA). For the current input $\bm{x}$, an aggregated input influence matrix $\bm{A}$ is constructed. This matrix $\bm{A}$ accumulates the input $\bm{x}$, scaled by its similarity weights $w_m$, into components corresponding to the memory keys that were among the top-$k$ selected. 
Each memory key $\bm{k}_m^{(t-1)}$ is then updated to $\bm{k}_m^{(t)}$ using $m$-th component of A, denoted $(\bm{A})_m$ 
\begin{equation}
    \bm{k}_{m}^{(t)} = (1 - \alpha) \cdot \bm{k}_{m}^{(t-1)} + \alpha \cdot (\bm{A})_m, \label{eq:mrffn_key_update}
\end{equation}
where $\bm{A_m}$ represents the accumulated information relevant to the $m$-th memory key, and $\alpha$ is the momentum coefficient (e.g., 0.1). This update mechanism adjusts the memory keys to better reflect the characteristics of input queries that frequently access them. The memory values $\bm{v_m}$ are learnable parameters updated via standard backpropagation.

The retrieved embedding $\bm{z_r}$ and the original input $\bm{x}$ are then concatenated and passed to a gating network within the MoE layer. This network computes probabilities $g_e$ for selecting specialized experts. The final output of MRFFN is a weighted combination of these selected expert outputs given by:
\begin{equation}
    \mathrm{out} = \sum_{e \in \text{top-}k} g_{e} \cdot \mathrm{Expert}_e(x), \quad  \bm{g} = \mathrm{softmax}(\bm{W_g} [\bm{x}; \bm{z_r}; \bm{\bar{a}}])
    \label{eq:mrffn_moe_output}
\end{equation}
Here, $\bm{W_g}$ is the gate's learnable weight matrix, and $\bm{\bar{a}}$ corresponds to an attention summary. This summary is derived from an internal attention mechanism where hidden states specific to each expert (provided as input to the MRFFN) query the MRFFN's memory module. This dynamic routing to experts based on memory retrieval allows MRFFN to capture complex temporal dependencies, enhancing forecasting stability and generalization for complex spatio-temporal patterns.


\begin{table*}[ht]
\centering
\caption{Model performance comparison on demand forecasting dataset: NYCTaxi, Citi Bike}
\label{tab:dataset_comparison}
\resizebox{\textwidth}{!}{
\begin{tabular}{lccc|ccc|ccc|ccc}
\toprule
\toprule
 \textbf{Dataset} & \multicolumn{3}{c|}{\textbf{NYCTaxi Pick-up}} & \multicolumn{3}{c|}{\textbf{NYCTaxi Drop-off}} & \multicolumn{3}{c|}{\textbf{Citi Bike Pick-up}} & \multicolumn{3}{c}{\textbf{Citi Bike Drop-off}} \\
\cmidrule(lr){0-1} \cmidrule(lr){2-4} \cmidrule(lr){5-7} \cmidrule(lr){8-10} \cmidrule(lr){11-13}
\textbf{Method} & \textbf{MAE} & \textbf{RMSE} & \textbf{MAPE} & \textbf{MAE} & \textbf{RMSE} & \textbf{MAPE} & \textbf{MAE} & \textbf{RMSE} & \textbf{MAPE} & \textbf{MAE} & \textbf{RMSE} & \textbf{MAPE} \\
\midrule
DCRNN & 5.40 & 9.71 & 35.09\% & 5.19 & 9.63 & 37.78\% & 2.09 & 3.30 & 54.22\% & 1.96 & 2.94 & 51.42\% \\
STGCN & 5.71 & 10.22 & 36.51\% & 5.38 & 9.60 & 39.12\% & 2.08 & 3.31 & 53.63\% & 2.01 & 3.07 & 50.45\% \\
ASTGCN & 7.43 & 13.84 & 47.96\% & 6.98 & 14.70 & 45.48\% & 2.76 & 4.45 & 64.23\% & 2.79 & 4.20 & 69.88\% \\
GWN & 5.43 & 9.39 & 37.79\% & 5.03 & 8.78 & 35.63\% & 2.04 & 3.20 & 53.08\% & 1.95 & 2.98 & 50.30\% \\
AGCRN & 5.79 & 10.11 & 40.40\% & 5.45 & 9.56 & 40.67\% & 2.16 & 3.46 & 56.35\% & 2.06 & 3.19 & 51.91\% \\
GMAN & 5.43 & 9.47 & 34.39\% & 5.09 & 8.95 & 35.00\% & 2.20 & 3.35 & 57.34\% & 2.09 & 3.00 & 54.82\% \\
STSGCN & 6.19 & 11.14 & 39.67\% & 5.62 & 10.21 & 37.92\% & 2.36 & 3.73 & 58.17\% & 2.73 & 4.50 & 57.89\% \\
ASTGNN & 5.90 & 10.71 & 40.15\% & 6.28 & 12.00 & 49.78\% & 2.37 & 3.67 & 60.08\% & 2.24 & 3.35 & 57.21\% \\
STG-NCDE & 6.24 & 11.25 & 43.20\% & 5.38 & 9.74 & 40.45\% & 2.15 & 3.97 & 55.49\% & 2.28 & 3.42 & 60.96\% \\
DGCRN & 5.44 & 9.82 & 35.78\% & 5.14 & 9.39 & 35.09\% & 2.06 & 3.21 & 54.06\% & 1.96 & 2.93 & 51.99\% \\
D$^{2}$STGNN & 5.32 & 9.12 & 35.51\% & 5.01 & 8.74 & 35.81\% & 2.02 & 3.18 &  53.60\% & 1.92 & 2.90 & 51.94\% \\
MegaCRN & 5.47 & 9.96 & 35.13\% & 5.07 & 9.11 & 35.08\% & 2.31 & 3.59 & 67.07\% & 2.18 & 3.30 & 61.42\% \\
STIDGCN & 5.15 & 8.90 & 33.74\% & 4.89 & 8.51 & 34.30\% & 2.00 & 3.11 & \textbf{51.71\%} & 1.88 & 2.80 & 49.43\% \\
OFA & 5.82 & 10.42 & 36.67\% & 5.60 & 10.14 & 37.39\% & 2.06 & 3.21 & 53.55\% & 1.96 & 2.97 & 49.64\% \\
GATGPT & 5.92 & 10.55 & 37.83\% & 5.66 & 10.39 & 39.36\% & 2.07 & 3.23 & 52.54\% & 1.95 & 2.94 & 49.26\% \\
ST-LLM & 5.29 & 9.42 & 33.55\% & 5.07 & 9.07 & 33.34\% & 1.99 & 3.08 & 53.54\% & 1.89 & 2.81 & 49.50\% \\
ST-LINK & \textbf{5.09} & \textbf{8.81} & \textbf{32.68\%} & \textbf{4.81} & \textbf{8.43} & \textbf{33.29\%} & \textbf{1.98} & \textbf{3.08} & 53.84\% & \textbf{1.87} & \textbf{2.80} & \textbf{49.23}\% \\
\bottomrule
\bottomrule
\end{tabular}
}
\end{table*}


\subsubsection{\textbf{Partially Frozen Attention LLM}}

Fine-tuning all layers of a pre-trained language model for spatio-temporal tasks can be computationally inefficient and prone to overfitting, particularly with limited traffic forecasting data. 
To address this, we adopt Partially Frozen Attention (PFA) LLM~\cite{stllm,frozenblip2}, selectively updating key model components to balance efficiency and adaptation while leveraging pre-trained knowledge for domain-specific learning with minimal fine-tuning. 
Unlike traditional methods that freeze feed-forward network (FFN) layers, our approach freezes the lower \(L - U\) layers entirely, including both multi-head attention (MHA) and Mixture-of-Experts (MoE)-based FFN, except for LayerNorm which remains trainable for stability. 
Conversely, only the upper \(U\) layers remain trainable, allowing MHA to refine spatio-temporal dependencies while the MoE-based FFN updates its gating mechanism and memory module. 
This targeted fine-tuning strategy ensures that pre-trained general knowledge is retained in the early layers, while later layers effectively adapt to traffic prediction.


\section{Evaluation}
\subsection{Experimental Settings}
\textbf{Datasets. } This study evaluates the performance of the proposed ST-LINK model on two spatio-temporal forecasting tasks: traffic demand prediction and traffic speed forecasting. For demand prediction, we utilize the NYCTaxi and Citi Bike datasets ~\cite{nyctaxibike}, where taxi pick-ups and bike rentals are aggregated at the station level. The NYCTaxi dataset consists of 266 virtual stations, while the Citi Bike dataset focuses on 250 stations. 
For traffic speed forecasting, we adopt widely used benchmark datasets~\cite{dcrnn}, including METR-LA and PEMS-BAY. METR-LA includes data from 207 sensors on Los Angeles highways, while PEMS-BAY consists of data from 325 sensors in the Bay Area. These datasets provide traffic sensor readings collected at 5-minute intervals, with variations in the number of sensors and time periods covered. For both demand and traffic speed forecasting tasks, we use historical observations from the past one hour (12 time steps) as input to predict the subsequent one hour (12 time steps), following standard spatio-temporal forecasting setups~\cite{dcrnn}.
\\
\\
\textbf{Baselines. } We compare \toolname with multiple baseline models across different forecasting tasks to comprehensively evaluate its effectiveness. For the demand forecasting task, we evaluate its performance against various deep learning-based spatio-temporal models, including RNN-based models DCRNN \cite{dcrnn}, AGCRN \cite{agcrn}, and DGCRN \cite{dgcrn}; graph and GCN-based models STGCN \cite{stgcn}, GWN \cite{gwn} STG-NCDE \cite{stg_ncde}, MegaCRN~\cite{megacrn}, D$^{2}$STGNN~\cite{D2STGNN} and STIDGCN~\cite{STIDGCN}; and attention-based models GMAN \cite{gman}, ASTGNN \cite{astgnn}, and ASTGCN \cite{astgcn}. Additionally, we assess its performance against LLM-based approaches, including OFA \cite{GPT4TS}, GATGPT \cite{gatgpt}, and ST-LLM \cite{stllm}. 
For the traffic speed forecasting task, we evaluate \toolname against LLM-based approaches: OFA \cite{GPT4TS}, GATGPT \cite{gatgpt}, and ST-LLM \cite{stllm}. 
We also provide case studies, offering a comparative analysis of these LLM methodologies in the traffic forecasting.
\\
\\
\textbf{Evaluation Setting. } To evaluate model \toolname, we employ multiple error metrics, including mean absolute error (MAE), root mean squared error (RMSE), and mean absolute percentage error (MAPE). 
We follow standard data splits for training, validation, and testing, applying a 6:2:2 ratio to the NYCTaxi and Citi Bike datasets and a 7:1:2 ratio to the METR-LA and PEMS-BAY datasets. 
All experiments are conducted on NVIDIA A6000 GPUs, using GPT-2~\cite{GPT2} as a backbone model. 
The batch size of 64 is used across all datasets.
We set the learning and drop out rates to 0.0001 and 0.1, respectively for the NYCTaxi and Citi Bike datasets, and 0.00025 and 0.3 for METR-LA and PEMS-BAY datasets. 
We have run each experiment five times and reported average performance.

\subsection{Experimental Results}
\subsubsection{\textbf{Demand Forecasting. }}
In the demand forecasting experiment, RNN-based models (DCRNN, AGCRN, DGCRN) leverage graph structures to model inter-node interactions, effectively capturing temporal dependencies. 
GWN, which utilizes a learnable graph structure, and D$^{2}$STGNN also yield competitive results. 
GCN-based models (STGCN, ASTGCN, MegaCRN, and STIDGCN) effectively capture spatial relationships but can encounter GCN-specific architectural limitations. 
Regarding model-specific comparisons, while STIDGCN achieves a better result than ST-LINK on the Citi Bike pickup MAPE metric, this particular outcome is influenced by that dataset's high proportion of zero targets (over 30\%). 
We see that \toolname demonstrates overall superior performance across the majority of other metrics and datasets.
Among attention-based models, GMAN effectively incorporates multi-attention mechanisms into graph structures, while ASTGNN and ASTGCN encounter difficulties in seamlessly integrating graph structures. 

For LLM-based models, GATGPT, which integrates Graph Attention with LLMs , exhibits improvements over OFA but still falls short in comparison to other architectures, suggesting that an optimal design that aligns model architecture with spatio-temporal characteristics is critical. ST-LLM, which combines a large language model with spatio-temporal embeddings, produces promising outcomes but struggles with the more complex spatial dataset NYCTaxi.

\subsubsection{\textbf{Traffic Forecasting. }}
We evaluate the performance of ST-LINK on traffic speed forecasting benchmarks, METR-LA (Table~\ref{tab:horizon_comparison1}) and PEMS-BAY (Table~\ref{tab:horizon_comparison2}). 
Experimental results demonstrate that \toolname consistently outperforms existing LLM-based baselines, including OFA, GATGPT, and ST-LLM. 
Notably, ST-LINK achieves the lowest error rates across all prediction horizons (3, 6, and 12) and evaluation metrics, establishing its effectiveness in both short and long-term forecasting. 
While ST-LLM yields comparable results in short-term forecasts, ST-LINK demonstrates a marked advantage at longer horizons (e.g., Horizon@12). 
This superior overall performance is attributed to ST-LINK's architectural design, where its SE-Attention effectively captures complex spatial dependencies, and its MRFFN enhances long-range temporal modeling by utilizing extended historical patterns.

\begin{table}[t]
\caption{Comparison of model performance across different forecasting horizons on the METR-LA dataset.}
\label{tab:horizon_comparison1}
\centering
\adjustbox{max width=\columnwidth}{
\begin{tabular}{@{}lccccc@{}}
\toprule
\textbf{Method} & \textbf{OFA} & \textbf{GATGPT} & \textbf{ST-LLM} & \textbf{ST-LINK} \\
\midrule
Horizon@3 MAE  & 2.83 & 2.85 & 2.82 & \textbf{2.77} \\
Horizon@3 RMSE & 5.51 & 5.54 & 5.54 & \textbf{5.37} \\
Horizon@3 MAPE & 7.34\% & 7.31\% & 7.31\%  & \textbf{7.23\%} \\
\midrule
Horizon@6 MAE  & 3.22 & 3.21 & 3.16 & \textbf{3.12} \\
Horizon@6 RMSE & 6.61 & 6.56 & 6.52 & \textbf{6.33} \\
Horizon@6 MAPE & 8.80\% & 8.74\%  & 8.68\% & \textbf{8.56\%} \\
\midrule
Horizon@12 MAE  & 3.66 & 3.66 & 3.58 & \textbf{3.51} \\
Horizon@12 RMSE & 7.68 & 7.63 & 7.53 & \textbf{7.27} \\
Horizon@12 MAPE & 10.44\% & 10.37\% & 10.29\% & \textbf{10.02\%} \\
\bottomrule
\end{tabular}
}
\end{table}

\begin{figure}[t] 
    \centering
    \includegraphics[width=\columnwidth]{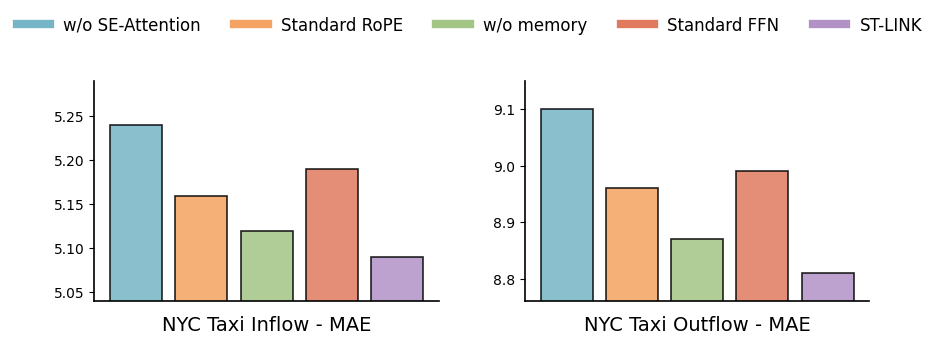} 
    \caption{Ablation study results of ST-LINK.}
    \label{fig:ablation}
\end{figure}

\subsection{Ablation Study}
We conducted an ablation study to verify the effectiveness of ST-LINK's key components:
\begin{itemize}[leftmargin=*, itemsep = 3pt, topsep = 10pt]
    \item \textbf{w/o SE-Attention. } The SE-Attention module preceding the LLM block is excluded to examine the effectiveness of SE-Attention in capturing spatio-temporal dependencies.
    \item \textbf{Standard RoPE. } The spatial extension of RoPE used in SE-Attention is excluded, while the standard RoPE employed in conventional LLMs is retained, allowing only sequential embedding without explicit spatio-temporal embeding.
    \item \textbf{w/o Memory. } The memory retrieval mechanism and the memory-expert attention computation in the MRFFN are excluded, forcing the model to operate without memory-based information retrieval. This experiment assesses the contribution of memory retrieval in searching for spatio-temporal patterns and enhancing the MoE gating network.
    \item \textbf{Standard FFN. } Instead of utilizing the MRFFN, the standard feed-forward network from the original LLM block is applied. This experiment measures the model's performance in the absence of memory-based retrieval and retrieval embedding-driven MoE gating.
\end{itemize}

Figure \ref{fig:ablation} presents the results conducted on the NYCTaxi dataset, showing each component's impact on ST-LINK performance. Removing SE-Attention notably decreases performance, confirming the importance of explicit spatial encoding. Similarly, replacing spatially extended RoPE with standard RoPE leads to performance degradation, highlighting spatial encoding’s critical role. The w/o memory experiment also shows reduced performance, indicating memory retrieval significantly improves forecasting accuracy through effective spatio-temporal pattern integration. Replacing MRFFN with a standard FFN further reduces performance, emphasizing the importance of memory retrieval-driven MoE gating for spatio-temporal learning. Overall, results confirm SE-Attention and MRFFN as essential, synergistic components of ST-LINK.

\begin{table}[t]
\caption{Comparison of model performance across different forecasting horizons on the PEMS-BAY dataset.}
\label{tab:horizon_comparison2}
\centering
\adjustbox{max width=\columnwidth}{
\begin{tabular}{@{}lccccc@{}}
\toprule
\textbf{Method} & \textbf{OFA} & \textbf{GATGPT} & \textbf{ST-LLM} & \textbf{ST-LINK} \\
\midrule
Horizon@3 MAE  & 1.37 & 1.38 & 1.33 & \textbf{1.32} \\
Horizon@3 RMSE & 2.88 & 2.87 & 2.81 & \textbf{2.77} \\
Horizon@3 MAPE & 2.87\% & 2.89\% & 2.80\% & \textbf{2.77\%} \\
\midrule
Horizon@6 MAE  & 1.72 & 1.74 & 1.67 & \textbf{1.63} \\
Horizon@6 RMSE & 3.90 & 3.88 & 3.80 & \textbf{3.68} \\
Horizon@6 MAPE & 3.82\% & 3.85\% & 3.72\% & \textbf{3.69\%} \\
\midrule
Horizon@12 MAE  & 2.04 & 2.06 & 1.95 & \textbf{1.90} \\
Horizon@12 RMSE & 4.69 & 4.69 & 4.49 & \textbf{4.34} \\
Horizon@12 MAPE & 4.63\% & 4.71\% & 4.53\% & \textbf{4.47\%} \\
\bottomrule
\end{tabular}
}
\end{table}

\begin{table*}[t]
    \centering
    \caption{Forecasting performance of ST-LINK against baseline models under defined scenarios: (I) Isolated Roads, (H) Hard-to-predict Roads, and (E) Time with Sudden Events.}
    \label{tab:IHE_table}
    \resizebox{\textwidth}{!}{
    \begin{tabular}{l | c c c | c c c | c c c | c c c | c c c | c c c}
        \toprule
        \textbf{Method} & \multicolumn{3}{c|}{\textbf{PEMS-BAY(I)}} & \multicolumn{3}{c|}{\textbf{PEMS-BAY(H)}} & \multicolumn{3}{c|}{\textbf{PEMS-BAY(E)}} & \multicolumn{3}{c|}{\textbf{METR-LA(I)}} & \multicolumn{3}{c|}{\textbf{METR-LA(H)}} & \multicolumn{3}{c}{\textbf{METR-LA(E)}} \\
        \cmidrule(lr){2-4} \cmidrule(lr){5-7} \cmidrule(lr){8-10} \cmidrule(lr){11-13} \cmidrule(lr){14-16} \cmidrule(lr){17-19}
        & MAE & RMSE & MAPE & MAE & RMSE & MAPE & MAE & RMSE & MAPE & MAE & RMSE & MAPE & MAE & RMSE & MAPE & MAE & RMSE & MAPE \\
        \midrule
        OFA     & 1.61 & 3.71 & 3.56\% & 1.66 & 3.81 & 3.66\% & 2.78 & 5.68 & 7.10\% & 3.18 & 6.39 & 8.64\% & 3.17 & 6.50 & 8.53\% & \textbf{3.78} & 7.65 & 11.06\% \\
        GATGPT  & 1.62 & 3.69 & 3.59\% & 1.67 & 3.80 & 3.70\% & 2.82 & 5.75 & 7.17\% & 3.18 & 6.37 & 8.61\% & 3.18 & 6.58 & 9.10\% & \textbf{3.78} & \textbf{7.63} & 11.04\% \\
        ST-LLM   & 1.55 & 3.58 & 3.47\% & 1.59 & 3.65 & 3.56\% & 2.68 & 5.47 & 6.92\% & 3.11 & 6.27 & 8.45\% & 3.11 & 6.41 & 8.46\% & 3.81 & 7.91 & 11.21\%     \\
        ST-LINK  & \textbf{1.52} & \textbf{3.49} & \textbf{3.46\%} & \textbf{1.55} & \textbf{3.55} & \textbf{3.50\%} & \textbf{2.61} & \textbf{5.25} & \textbf{6.76\%} & \textbf{3.07} & \textbf{6.15} & \textbf{8.43\%} & \textbf{3.05} & \textbf{6.21} & \textbf{8.28\%} &  3.79 &  7.73 &  \textbf{10.93\%} \\
        \bottomrule
    \end{tabular}
    }
\end{table*}

\begin{figure*}[t] 
    \centering
    \includegraphics[width=\textwidth]{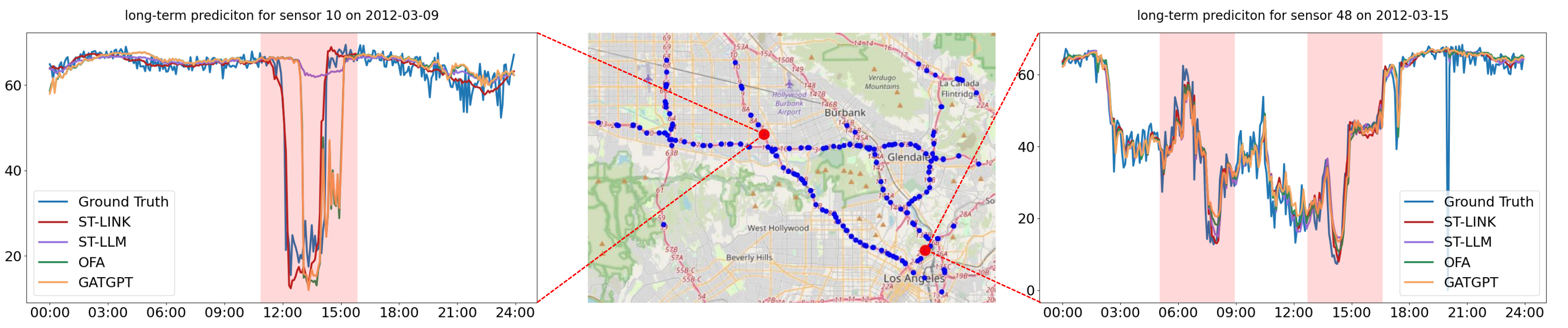} 
    \caption{Traffic forecasting case study for two sensors in METR-LA.}
    \label{fig:case1}
\end{figure*}

\subsection{Case Study}
\subsubsection{\textbf{Case-specific Experiments}}
Previous studies in traffic forecasting have evaluated model performance under various fluctuating conditions that can occur in real-world road environments. 
In this study, to verify the robustness and practical predictive capability of the proposed model, we adopt three specific experimental scenarios originally outlined by prior research~\cite{testam}. 
These scenarios are (I) Isolated Roads, (H) Hard-to-predict Roads, and (E) Time with Sudden Events. 
Isolated roads (I) are segments with minimal connectivity to other roads, identified via topological analysis of the adjacency matrix targeting low-degree nodes. 
Hard-to-predict roads (H) include major intersections (identified by high node degrees) and segments with highly fluctuating traffic, selected based on top 10\% time-series entropy complexity. 
Time with sudden events (E) denotes periods of abrupt, network-wide traffic flow changes (e.g., accidents), identified by analyzing changes in a network mean state indicator derived from short-term traffic observations and selecting instances where change magnitudes are in the top 10\%.

Experimental results indicate the ST-LINK model generally performs well compared to baseline methods across most evaluation scenarios on both PEMS-BAY and METR-LA datasets. However, in the METR-LA (E) scenario, which represents sudden network-wide traffic events, the OFA and GATGPT models also show good performance. Overall, ST-LINK delivers robust prediction accuracy and demonstrates consistent performance in locations with high variability and during abrupt traffic changes.

\subsubsection{\textbf{Qualitative Evaluation}}
Figure~\ref{fig:case1} presents a case study from the METR-LA dataset, focusing on long-term traffic speed forecasting for two selected sensors: Sensor 10 (March 9, 2012) and Sensor 48 (March 15, 2012). 
This case study examines how ST-LINK and baseline models capture abrupt changes and fluctuations in traffic speed.
In Figure~\ref{fig:case1} (left), the selected time window for Sensor 10 includes a sudden decrease in speed, which poses a challenge for predictive models. 
While most models fail to accurately predict the drop, ST-LINK aligns more closely with the ground truth, demonstrating improved performance in capturing the transition. 

Figure~\ref{fig:case1} (right) presents the forecasting results for Sensor 48, which exhibits frequent variations in traffic speed. The highlighted red boxes indicate periods of high volatility, during which ST-LINK provides a better match with the ground truth compared to other models. While baseline models follow the overall pattern, ST-LINK shows a distinct advantage in capturing both the overall trend and short-term fluctuations, particularly during abrupt changes. This capability is attributed to MRFFN, which dynamically retrieves and integrates past spatio-temporal patterns, allowing the model to adapt more effectively to complex variations. Overall, this case study demonstrates that ST-LINK effectively captures both sudden transitions and recurring patterns in traffic speed, highlighting its robustness in complex traffic forecasting tasks.

\section{Discussion}
The proposed MRFFN leverages a memory retrieval-based MoE framework to adapt to dynamic patterns. This design enhances the model’s ability to capture long-term dependencies and dynamic spatiotemporal patterns. However, several key aspects of its rapid adaptation process require further discussion.

One critical consideration is the impact of domain shifts and the model’s adaptation speed on learning stability. While the ability to quickly integrate new patterns is advantageous in dynamic environments, excessively rapid updates may lead to catastrophic forgetting~\cite{cf1,cf2}, where previously learned information is prematurely overwritten. Therefore, further analysis is required to understand how MRFFN’s memory management affects long-term learning, ensuring that the model remains adaptable to domain shifts while effectively retaining prior knowledge. Future research should explore refined memory update mechanisms to balance adaptation and stability.

In this study, ST-LINK is implemented using a GPT-2. Therefore, the proposed MRFFN and SE-Attention mechanisms are designed to be applicable to various LLMs and spatio-temporal forecasting models. While our experiments employed relatively constrained model sizes, it is essential to examine how these mechanisms scale to larger models.
In particular, analyzing the contributions of MRFFN and SE-Attention to spatio-temporal learning across different model sizes is crucial. LLMs may provide richer memory representations and better capture complex spatio-temporal dependencies, yet increased computational costs and memory requirements could limit their practicality. Therefore, it is necessary to assess whether these mechanisms consistently improve performance across diverse model sizes. Beyond simple scaling, developing strategies to maintain optimal performance across varying model capacities is essential.
\section{Conclusion}
In this paper, we propose \toolname, a novel framework designed to enhance spatio-temporal forecasting performance using large language models (LLMs). 
To address the challenges of applying LLMs to spatio-temporal data, we introduce RoPE-based SE-Attention, enabling effective learning of both temporal and spatial dependencies, and incorporate a Memory-Retrieval FFN (MRFFN) to preserve and utilize long-term time-series information. 
Experimental results demonstrate that \toolname outperforms or achieved comparable performance to existing forecasting models across various spatio-temporal benchmark datasets, validating the feasibility of leveraging LLMs for spatio-temporal forecasting tasks. 
These findings suggest that LLMs, beyond their traditional applications in text processing, can be effectively utilized in domains where spatio-temporal context is critical. 
Future research will focus on further generalizing \toolname to a wider range of spatio-temporal domains.

\begin{acks}
This work was supported by the National Research Foundation of Korea (NRF) grant funded by the Korea government (MSIT) (No. RS-2024–00456247, No. RS-2023–00218913) and by Institute of Information \& communications Technology Planning \& Evaluation (IITP) grant funded by the Korea government(MSIT) (No. RS-2025-25443718, Next-HCAI Project) and (No.RS-2019-II191906, Artificial Intelligence Graduate School Program(POSTECH).
\end{acks}

\appendix
\section*{Appendix}
\section{Reversible Instance Normalization}
\label{sec:revin}
Spatio-temporal data often exhibit node-specific scale discrepancies, which can lead to instability and biased attention. To address this, we normalize each node's temporal sequence individually using RevIN~\cite{revin}. 
In this context, for an input instance $x^{(i)}$ (from a set of $N$ sequences) which comprises $K$ variables (nodes) over an input sequence of length $T_x$, $x_{kj}^{(i)}$ denotes the value of the $k$-th variable at time step $j$ for the $i$-th instance. The instance-specific mean ${\mathbb{E}t}[x_{kt}^{(i)}]$ and variance ${Var}[x_{kt}^{(i)}]$ for each $k$-th variable's temporal sequence (i.e., for $x_{k\cdot}^{(i)} \in \mathbb{R}^{T_x}$) are computed as follows:
\begin{equation*}
{\mathbb{E}t}[x_{kt}^{(i)}] = \frac{1}{T_x} \sum_{j=1}^{T_x} x_{kj}^{(i)}, \ \ \ {Var}[x_{kt}^{(i)}] = \frac{1}{T_x} \sum_{j=1}^{T_x} \left(x_{kj}^{(i)} - \operatorname{\mathbb{E}t}[x_{kt}^{(i)}]\right)^2
\end{equation*}
Using these statistics, we normalize each input data point $x_{kt}^{(i)}$ from the instance $x^{(i)}$ as:
\begin{equation*}
\hat{x}_{kt}^{(i)} = \gamma_k \left( \frac{x_{kt}^{(i)} - {\mathbb{E}t}[x_{kt}^{(i)}]}{\sqrt{{Var}[x_{kt}^{(i)}] + \epsilon}} \right) + \beta_k,
\end{equation*}
where $\gamma_k$ and $\beta_k$ are the $k$-th components of learnable affine parameter vectors $\gamma, \beta \in \mathbb{R}^K$, and $\epsilon$ is a small constant added for numerical stability. 
This instance-specific normalization, using the mean and standard deviation computed over the temporal dimension for each node, ensures that attention mechanisms can compare features across different nodes more fairly, without being distorted by differences in their magnitudes or statistical properties.

\section{Efficiency Study}
To evaluate the efficiency of ST-LINK, we compare its model size, number of trainable parameters, training time, and inference speed against baseline LLM models, as summarized in Table ~\ref{tab:model_comparison}. ST-LINK significantly reduces the number of parameters while maintaining effective spatio-temporal learning. Unlike existing models that rely on large-scale parameters, ST-LINK enhances computational efficiency through memory-based dynamic gating within MRFFN, which incorporates a Mixture of Experts (MoE) mechanism. MRFFN dynamically retrieves and integrates past spatio-temporal patterns, minimizing redundant computations, while MoE selectively activates expert networks based on specific temporal and spatial characteristics, allowing for targeted processing. This mechanism not only optimizes model utilization but also contributes to faster inference by reducing unnecessary computations and focusing resources on relevant patterns. In terms of computational cost, ST-LINK reduces training time by approximately 38.2\% compared to ST-LLM, while inference speed improves by about 37.5\%. These efficiency gains demonstrate ST-LINK's ability to streamline spatio-temporal forecasting while maintaining high predictive performance.
\begin{table}[]
    \caption{Computational efficiency analysis of different models, comparing parameter count, trainable parameters, training time per epoch, and inference speed.}
    \label{tab:model_comparison}
    \centering
    \adjustbox{max width=\columnwidth}{
    \begin{tabular}{lccccc}
        \toprule
        dataset & \#parameter & \#trainable & Train Time/epoch & Inference Time \\
        \midrule
        OFA & 81M & 0.8M & 14.7373 secs & 2.2637 secs \\
        GATGPT & 82M & 1.7M & 14.5591 secs & 2.2071 secs \\
        ST-LLM & 82M & 42M & 15.1122 secs & 2.2231 secs \\
        ST-LINK & 10M & 2.2M & 9.3425 secs & 1.3899 secs \\
        \bottomrule
    \end{tabular}
    }
\end{table}

\section{Algorithm}
\begin{algorithm}
\caption{ST-LINK}
\label{alg:st_link_concise}
\begin{spacing}{1.1}
\begin{algorithmic}[1]
\Statex \textbf{Input:} Historical spatio-temporal data $X_{hist}$ ($B \times T_{in} \times N \times F_{in}$), embedded via convolution and fused with temporal and node embeddings into $H^{(0)}$.
\Statex
\Statex \textbf{Model ($L$ layers):}
\State \quad Each layer $l$ processes $H^{(l-1)}$ (or $H^{(0)}$) via two blocks:
\Statex \quad \textbf{a. SE-Attention:}
\State \quad Normalize input $H^{(l-1)}$ with RevIN:
\Statex \quad\quad\quad $H_{\text{norm}}^{(l-1)} = \text{RevIN}(H^{(l-1)})$
\State \quad Project normalized input into Q, K, and V.
\State \quad Enhance Q, K via RoPE$_T$ and RoPE$_S$ embeddings:
\Statex \quad\quad\quad $\Phi(Q_i) = W_q \left[\text{RoPE}_T(Q_i; t_i); \text{RoPE}_S(Q_i; N_{in})\right]$,
\Statex \quad\quad\quad $\Phi(K_i) = W_k \left[\text{RoPE}_T(K_i; t_i); \text{RoPE}_S(K_i; N_{in})\right]$
\State \quad Compute scaled dot-product attention:
\Statex \quad\quad\quad $Attn(Q,K,V)_i = \sum_j \frac{\exp\left(\Phi(q_i)^\top \Phi(k_j)/\sqrt{d_h}\right)}{\sum_n \exp\left(\Phi(q_i)^\top \Phi(k_n)/\sqrt{d_h}\right)} v_j$
\State \quad Denormalize (RevIN)
\State \quad Add residual connection, and apply LayerNorm.
\State \quad The output of this block : ${H}_{\text{attn}}$
\Statex \quad \textbf{b. MRFFN:}
\State \quad Retrieve Historical Pattern ${z_r} = \sum_{m \in \text{top-}k} w_m {v_m}$
\State \quad Update Memory Keys ${k_m}$ (via EMA):
\Statex \quad\quad\quad ${k}_{m}^{(t)} = (1 - \alpha) \cdot {k}_{m}^{(t-1)} + \alpha \cdot {A}_m$
\State \quad Compute Attention Summary ${\bar{a}}$
\State \quad Final Output :
\Statex \quad\quad\quad $\text{MRFFN}_{\text{Output}} = \sum_{e \in \text{top-}k} g_{e} \cdot \text{Expert}_e({H}_{\text{attn}})$
\Statex \quad\quad\quad where ${g} = \text{softmax}({W_g} [{H}_{\text{attn}}; {z_r}; {\bar{a}}])$
\State \quad Apply Residual Connection:
\Statex \quad\quad\quad ${H}^{(l)} = \text{LayerNorm}(\text{MRFFN}_{\text{Output}} + {H}_{\text{attn}})$

\Statex
\Statex \textbf{Prediction Head:}
\State Project final encoder output $H^{(l)}$ to prediction:
\Statex \quad \quad\quad $Y_{\text{pred}} = \text{Regression}(H^{(l)}) \in \mathbb{R}^{B \times T_{\text{out}} \times N \times F_{\text{out}}}$

\Statex \textbf{Return} $Y_{\text{pred}}$.
\end{algorithmic}
\end{spacing}
\end{algorithm}

\newpage




\section*{GenAI Usage Disclosure}
During the preparation of this paper, we used generative AI only for grammar and style checking in English.
No generative AI was used in the conceptualization of the research, design of experiments or writing the main content of the paper.

\bibliographystyle{ACM-Reference-Format}
\balance
\bibliography{sample-base}

\end{document}